\documentclass[letterpaper, 10 pt, conference]{ieeeconf}
\IEEEoverridecommandlockouts
\usepackage{amsmath,amsfonts}
\usepackage{array}
\usepackage{stfloats}
\usepackage[hyphens]{url}
\usepackage{graphicx}
\usepackage{cite}
\usepackage{tabularx}
\usepackage[hidelinks]{hyperref}
\newcolumntype{C}{>{\centering\arraybackslash}X}

\begin{document}

\title{\LARGE \bf Design and Validation of a Robotic Knee Exoskeleton with Musculoskeletal Simulation-Derived Biological Torque Control}

\author{Rajiv V. Joshi$^{1}$, Jimin An$^{1}$, Ilseung Park$^{1}$, Dawit Lee$^{2}$, and Inseung Kang$^{1}$
\thanks{This work has been submitted to the IEEE for possible publication. Copyright may be transferred without notice, after which this version may no longer be accessible.}
\thanks{$^{1}$R. Joshi, J. An, I. Park, and I. Kang are with the Department of Mechanical Engineering, Carnegie Mellon University, Pittsburgh, PA, 15213 USA (e-mail: {\tt\footnotesize rvj@andrew.cmu.edu}). $^{2}$D. Lee is with the Department of Bioengineering, Stanford University, Stanford, CA, 94305}
}

\maketitle
\thispagestyle{empty}
\pagestyle{empty}

\begin{abstract}
Data-driven controllers have enabled lower-limb robotic exoskeletons to adapt assistance to the user across many tasks. A leading example is biological torque control, which estimates the user's own joint moment from onboard sensors and assists in proportion to that estimate. However, its estimators are trained on data recorded with the target device, so deploying the controller on new hardware requires collecting new data. Existing open-source efforts release hardware, software, or controllers separately, so this data collection must still be repeated on the assembled device. Here, we present an open-source knee exoskeleton, built from off-the-shelf components and widely available fabrication tools, with a biological torque controller derived entirely from simulation. A muscle-driven imitation policy reproduces squatting motions from an open biomechanics dataset, and the resulting knee moments are distilled into an estimator that uses only onboard sensors, so no training data are recorded on the device. In seven able-bodied participants, assistance reduced peak biological knee extension moment by 21.5\% and peak vastus medialis activation by 26.7\% during squatting, relative to squatting without the exoskeleton (p$<$0.05). Our results demonstrate that a biological torque controller derived from simulation can transfer to physical hardware without device-specific data collection. Because the hardware and training pipeline are released together, research groups can train and compare knee exoskeleton controllers on shared hardware without additional data collection overhead.
\end{abstract}


\begin{figure*}[!t]
    \centering
    \includegraphics[width=\textwidth]{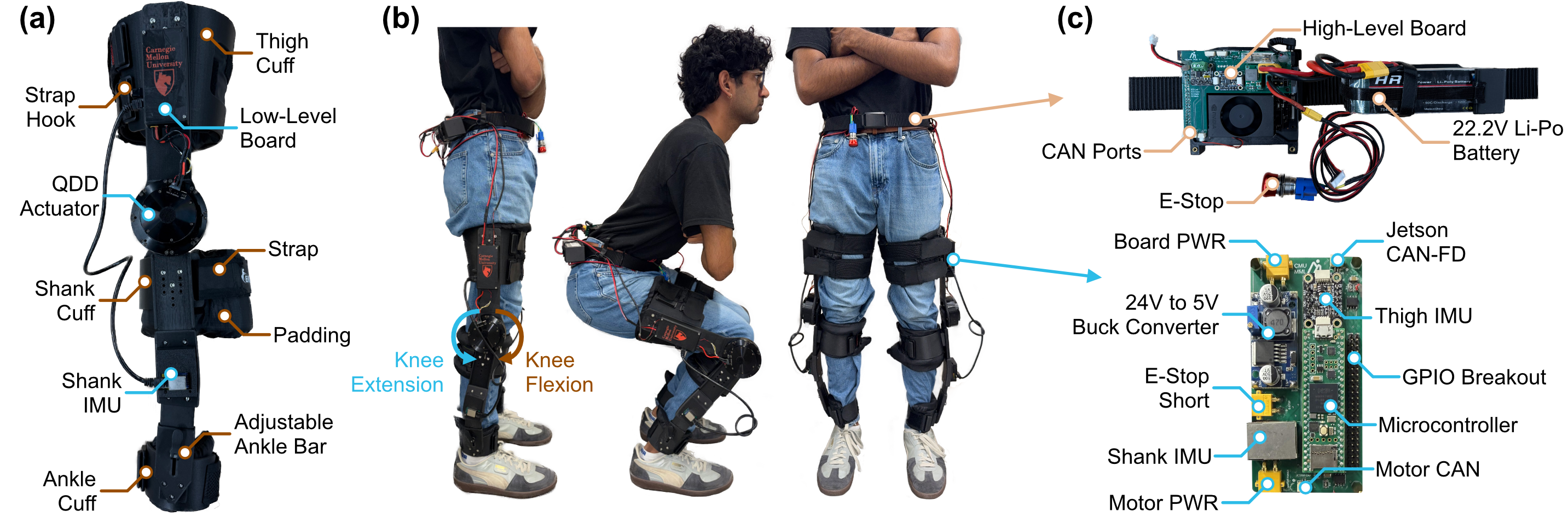}
    \caption{
    \textbf{Overview of the robotic knee exoskeleton design.} (a) Sagittal-plane view of the unilateral device, with straps, hooks, and cuffs at the thigh, shank, and ankle. (b) A participant wearing the bilateral device, shown left to right in sagittal standing, sagittal squatting, and frontal standing. (c) The high-level control board built around the embedded computer (top) and the low-level control board built around the microcontroller (bottom).
    }
    \label{fig:Exo_device}
\end{figure*}

\section{Introduction}
Wearable robotic exoskeletons have emerged as a promising intervention for people with motor impairments \cite{Sawicki2020TheEE, Young2017StateOfTheArt, Dollar2008Exoskeletons, Siviy2023Opportunities}. The knee is a particularly important target joint, bearing the majority of the load during high-demand activities such as squatting and stair negotiation \cite{Samuel2011StairDemand, Escamilla2001SquatBiomechanics}. Over the past decade, exoskeleton control has advanced rapidly, as data-driven methods now estimate gait phase and joint moments in real time from wearable sensors \cite{Kang2021GaitPhase, Molinaro2024JointMoment, Molinaro2024TaskAgnostic, An2024LocomotorTaskSets}, allowing assistance to adapt to the user's movement across tasks. Among these data-driven controllers, biological torque control is a leading approach, in which a neural network estimates the user's joint moment from onboard sensors and the exoskeleton applies a scaled, delayed copy of this estimate \cite{Molinaro2024JointMoment}. Because assistance follows the user's own effort rather than a predefined profile, the approach has reduced metabolic cost during level and inclined walking \cite{Molinaro2024JointMoment} and has been extended to unstructured locomotor tasks with a single controller \cite{Molinaro2024TaskAgnostic}.

However, this generality depends on device-specific data. The estimators are trained on large datasets recorded with the target device \cite{Molinaro2024TaskAgnostic} and tend to overfit to that device's sensing configuration \cite{Hwang2026UnifiedKinematic}. Collecting and labeling such data is time consuming, requires expensive laboratory equipment, and must be repeated for each new device \cite{Scherpereel2025DomainAdaptation}. Because these controllers are typically developed and validated on custom hardware built by the same research group \cite{Chen2021WearableKnee, Yu2025ControllingPower, Huang2022StiffnessBased, Boon2025StairAscent, Divekar2024VersatileKnee}, each controller also remains coupled to the device on which it was developed, which limits its reproducibility across research groups.

To lower this barrier, open-source efforts have released individual components of the exoskeleton system, ranging from modified commercial orthoses and mechanical designs to embedded software and trained controllers \cite{Nesler2022MBLUE, Lee2024AIUniversalExo, Leestma2025EpicallyPowerful, Molinaro2024TaskAgnostic}. These components are released separately and were not designed to operate together, so new data must still be collected on the assembled device before a data-driven controller can be deployed. In parallel, simulation and data reuse offer a path to reduce this data burden, including reinforcement learning in musculoskeletal simulation, adaptation of existing biomechanics datasets, and device-independent kinematic representations \cite{Park2026PredictiveHip, Choi2026MusculoskeletalImitation, Scherpereel2025DomainAdaptation, Hwang2026UnifiedKinematic}. However, a key limitation remains. No current exoskeleton approach has paired open-source hardware on which other groups could reproduce the full pipeline, from hardware to a simulation deployed controller.

We address these limitations by deriving a knee biological torque controller entirely from simulation and deploying it on an exoskeleton designed for accessible fabrication. Reference squatting motions from an open biomechanics dataset are reproduced by a muscle-driven imitation policy, and the knee moments generated in simulation are distilled into an estimator that uses only the exoskeleton's onboard sensors. Because no training data are recorded on the device, the same pipeline can be applied to other builds of the hardware. The exoskeleton is built from off-the-shelf components and parts fabricated with widely available tools, which minimizes the manufacturing overhead for other groups. During development, this construction also allowed the physical interface to be refined alongside the controller during development.

In this work, we present the following: 1) a knee exoskeleton, including its mechanical design, electronics, embedded firmware, and a two-tier control architecture; 2) a biological torque controller trained on an open biomechanics dataset through muscle-driven motion imitation and distilled to onboard sensors, requiring no data recorded on the device; and 3) an experimental validation of this controller in seven participants during squatting. We hypothesized that the controller would reduce knee extensor effort, quantified by peak biological knee extension moment and vastus medialis activation, relative to squatting without the exoskeleton. All hardware designs, code, and training pipelines are publicly available at: {\hypersetup{pdfborder={0 0 1},urlbordercolor={0 1 1}}\url{https://github.com/MetaMobilityLabCMU/Knee-Exoskeleton-Project}}.


\begin{table}[!t]
    \caption{Device Specifications\label{tab:specs}}
    \centering
    \renewcommand{\arraystretch}{1.2}
    \begin{tabularx}{\columnwidth}{l >{\raggedleft\arraybackslash}X}
    \hline
    \multicolumn{2}{c}{\textbf{Mechanical}}\\
    \hline
    Total mass, bilateral$^{*}$ (kg) & 3.5\\
    Unilateral mass (kg) & 1.3\\
    Actuator mass (kg) & 0.49\\
    Rated / peak torque$^{**}$ (Nm) & 9 / 22\\
    Range of motion (deg) & 0--120\\
    \hline
    \multicolumn{2}{c}{\textbf{Compute/Electrical}}\\
    \hline
    Inertial sensor communication & SPI\\
    Motor communication & CAN\\
    Low-to-High level communication & CANFD\\
    Low-level loop execution ($\mu$s) & $\sim$80\\
    High/low-level round trip (ms) & $<$0.5\\
    Estimated run time (h) & $\sim$2\\
    \hline
    \end{tabularx}
    \par\vspace{3pt}
    \begin{minipage}{\columnwidth}
    \footnotesize\raggedright
    $^{*}$Includes the battery and embedded electronics.\\
    $^{**}$The stock motor driver limits output to 18~Nm per the manufacturer's safety constraint.
    \end{minipage}
\end{table}

\section{Robotic Knee Exoskeleton Design}
\subsection{Mechanical Structure and User Interface}
The exoskeleton consists of two leg modules, each attached to the limb through three cuffs at the thigh, shank, and ankle (Fig.~\ref{fig:Exo_device}a; Table~\ref{tab:specs}). Because squatting places far greater extension than flexion demand on the knee \cite{Escamilla2001SquatBiomechanics}, the cuff layout prioritizes the transfer of extension torque. Each cuff wraps around the anterior and posterior surfaces of the limb, sharing the load between shear and normal force, which prevents the device from rotating about the limb without additional straps. This configuration was adopted after an earlier prototype, in which cuffs mounted only on the lateral side of the leg transmitted the extension torque purely through shear. The resulting transverse-plane moment rotated the device about the long axis of the limb at higher torques and corrupted the onboard sensor signals used by the controller. The ankle cuff serves as the distal anchor and limits migration of the device along the leg.

All cuffs follow the conical taper of the limb, adjust longitudinally to fit different segment lengths, and fasten with padded straps through integrated tabs. Beneath them, a two-layer liner of 3D mesh fabric over polyurethane foam distributes contact pressure and reduces pinching, skin irritation, and strap chafing. The cuffs and all other structural parts were printed from a carbon fiber reinforced polyamide (PAHT-CF, Bambu Lab, China), selected for its stiffness, strength, and heat resistance near the actuator. Print orientation and infill were chosen to favor shell-dominant bending stiffness, with solid outer walls and low infill density to minimize mass \cite{Markforged2026PrintStrength}.

\subsection{Actuation and Embedded System}
At the knee, torque is supplied by a quasi-direct-drive (QDD) actuator (AK80-9 V3.0, CubeMars, China) with an integrated motor driver and encoder (Table~\ref{tab:specs}). The actuator housing is fixed to the thigh bar and its output shaft drives the shank bar, so extension torque reaches the limb through the cuffs. The low gear ratio keeps the joint backdrivable, allowing the user to move with little impedance when no torque is commanded \cite{Huang2022StiffnessBased, Zhu2022Backdrivability}. This actuator has also been used in several other exoskeletons \cite{Nesler2022MBLUE, Lee2024AIUniversalExo, Zhao2024AnkleExo}, which simplifies replication with an off-the-shelf part.

Control of the actuator is split across two tiers. On each leg, a microcontroller (Teensy 4.1, PJRC, USA) runs a 1000~Hz low-level loop that reads two inertial measurement units (IMUs) on the thigh and shank (ICM-20948, TDK InvenSense, USA) and exchanges torque commands and motor states with the actuator. A waist-mounted embedded computer (Jetson Orin Nano, NVIDIA, USA) runs the 100~Hz high-level controller (Section~\ref{sec:controller}), matching the sampling rate of the laboratory instrumentation. In each cycle, the embedded computer polls both microcontrollers for their latest states and returns a torque command, and the round-trip time leaves most of the cycle for computing the controller output (Table~\ref{tab:specs}). For safety, a watchdog on each microcontroller commands zero torque if no valid command arrives within the expected cycle window, returning the joint to a passive, backdrivable state after any communication fault. A waist-mounted 22.2V lithium polymer battery (HRB, China) powers both tiers (Table~\ref{tab:specs}).


\begin{figure}[!t]
    \includegraphics[width=1\columnwidth]{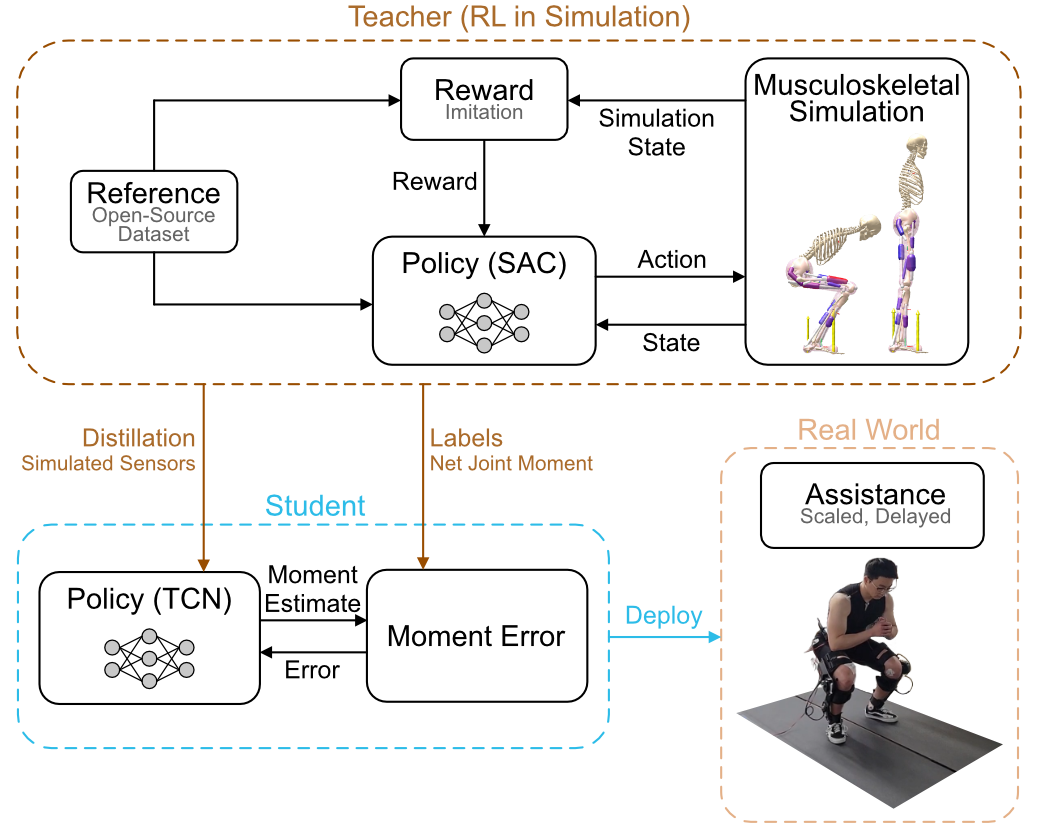}
    \caption{
    \textbf{Overview of the simulation-derived biological torque controller.} An imitation policy is trained with reinforcement learning in a musculoskeletal simulation to reproduce reference squatting motion, using an imitation reward as its learning signal. Knee muscle moments from the policy rollouts are then distilled into a temporal convolutional network that estimates the knee moment from onboard sensor signals, trained by minimizing the error against the simulated moments. On the exoskeleton, the estimated knee moment is scaled and delayed to generate assistive torque.
}
    \label{fig:Simulation}
\end{figure}

\section{Simulation-Derived Biological Torque Control}
\label{sec:controller}
The controller was derived through a two-stage framework that connects reference human kinematics to sensor-based knee assistance (Fig.~\ref{fig:Simulation}). In the first stage, a muscle-driven policy was trained to reproduce reference squatting kinematics in a physics-based musculoskeletal simulation, generating synchronized kinematics and knee joint moments. In the second stage, these simulated moments were distilled into an estimator that uses signals available from the onboard sensors (knee angle and the thigh and shank angular velocities). During deployment, the estimator output is scaled to generate assistive knee torque, following the biological torque control approach \cite{Molinaro2024JointMoment, Molinaro2024TaskAgnostic}. Because neither stage requires data recorded on the device, the resulting controller can be deployed on other builds of the hardware without additional data collection.

\subsection{Muscle-Driven Motion Imitation}
Reference squatting kinematics were taken from an open biomechanics dataset \cite{Scherpereel2023Dataset}, low-pass filtered with a sixth-order, zero-phase Butterworth filter at 4~Hz, and differentiated to obtain joint velocities. A sagittal-plane musculoskeletal model with seven degrees of freedom, nine muscles in a single lower limb, and a lumbar actuator was implemented in SCONE \cite{Geijtenbeek2019SCONE} with the Hyfydy physics engine \cite{Geijtenbeek2021Hyfydy}. Following the DeepMimic framework \cite{Peng2018DeepMimic}, a policy trained with soft actor--critic \cite{Haarnoja2018SoftActorCritic} reproduced the reference squat with this model. The policy observed joint and muscle states, foot contact, and upcoming reference kinematics, and output muscle excitations and a lumbar target angle, the latter tracked by a proportional--derivative controller.

The reward combined motion tracking with penalties on muscular effort and abrupt input changes, which encourage physiologically plausible muscle coordination and smooth joint moments for estimator training:
\begin{equation}
    r_t =
    w_p r_{p,t}
    + w_v r_{v,t}
    + w_e r_{e,t}
    - w_a c_{a,t}
    - w_u c_{u,t},
    \label{eq:imitation_reward}
\end{equation}
where $r_t$ is the reward at timestep $t$; $r_{p,t}$, $r_{v,t}$, and $r_{e,t}$ reward tracking of joint coordinates, joint velocities, and foot position relative to the pelvis, each formulated as an exponential function of the corresponding squared error; $c_{a,t}$ is the mean squared muscle activation; $c_{u,t}$ penalizes step-to-step changes in muscle excitation and normalized lumbar torque; and $w_p=0.60$, $w_v=0.15$, $w_e=0.25$, and $w_a=w_u=2.0$ are the corresponding weights. The policy ran at 100~Hz in episodes of up to 10~s, initialized from sampled reference states and terminated early if the pelvis deviated from the reference. Knee muscle moments from the resulting muscle-driven squats served as training targets for the onboard estimator.

\subsection{Distillation to Onboard Sensors}
The imitation policy relies on quantities the exoskeleton cannot measure, including muscle states, contact forces, and future reference motion. A separate estimator was therefore trained to predict the knee joint moment from sensor-equivalent signals recorded during deterministic rollouts of the policy. At each 100~Hz step, the estimator input is
\begin{equation}
    \mathbf{x}_t =
    \begin{bmatrix}
        \omega_{\mathrm{thigh},z,t} &
        \omega_{\mathrm{shank},z,t} &
        q_{\mathrm{knee},t}
    \end{bmatrix}^{\mathsf{T}},
    \label{eq:student_input}
\end{equation}
where $\omega_{\mathrm{thigh},z,t}$ and $\omega_{\mathrm{shank},z,t}$ are the sagittal-plane angular velocities of the thigh and shank, matching the onboard IMUs, and $q_{\mathrm{knee},t}$ is the knee angle, matching the actuator encoder. A temporal convolutional network \cite{Lea2016TemporalConvNet} maps a 95-sample window (0.95~s) of these inputs to the current moment estimate:
\begin{equation}
    \hat{y}_t = f_{\theta}\left(\mathbf{x}_{t-94},\ldots,\mathbf{x}_{t}\right),
    \label{eq:student_estimator}
\end{equation}
where $\hat{y}_t$ is the estimated knee moment, $f_{\theta}$ is the network with parameters $\theta$, and the training target $y_t$ is the simulated knee muscle moment scaled by a fixed normalization factor.

The network stacks residual blocks of dilated causal convolutions with a linear readout, so each estimate depends only on current and past samples, as required for real-time deployment. During training, inputs were standardized, and Gaussian noise on the gyroscope channels and offsets on the knee angle were added to expose the estimator to sensor noise and calibration error. The network was trained to minimize the mean squared error between $\hat{y}_t$ and $y_t$ using the AdamW optimizer with a learning rate of $5\times10^{-4}$ and a batch size of 64 for 50 epochs, and the checkpoint with the lowest validation loss was retained.

\subsection{Hardware Deployment}
The trained estimator runs on the embedded computer at 100~Hz using the onboard IMU and encoder signals. Following the biological torque control approach \cite{Molinaro2024JointMoment}, the commanded torque is a scaled, time-delayed copy of the estimated knee moment,
\begin{equation}
    \tau_{\mathrm{cmd}}(t) = k\,\hat{\tau}_{\mathrm{knee}}(t-\Delta t),
    \label{eq:biotorque}
\end{equation}
where $\tau_{\mathrm{cmd}}$ is the commanded knee torque, $\hat{\tau}_{\mathrm{knee}}$ is the estimator output $\hat{y}_t$ in (\ref{eq:student_estimator}) converted back to Nm, $k$ is the assistance gain, and $\Delta t$ is the assistance delay. The gain and delay were tuned for each participant during familiarization to set the magnitude and timing of assistance. The commanded torque was then clamped to 18~Nm, the output limit of the stock motor driver (Table~\ref{tab:specs}).


\section{Experimental Validation}
\subsection{Participants and Protocol}
\begin{figure}[!t]
    \includegraphics[width=1\columnwidth]{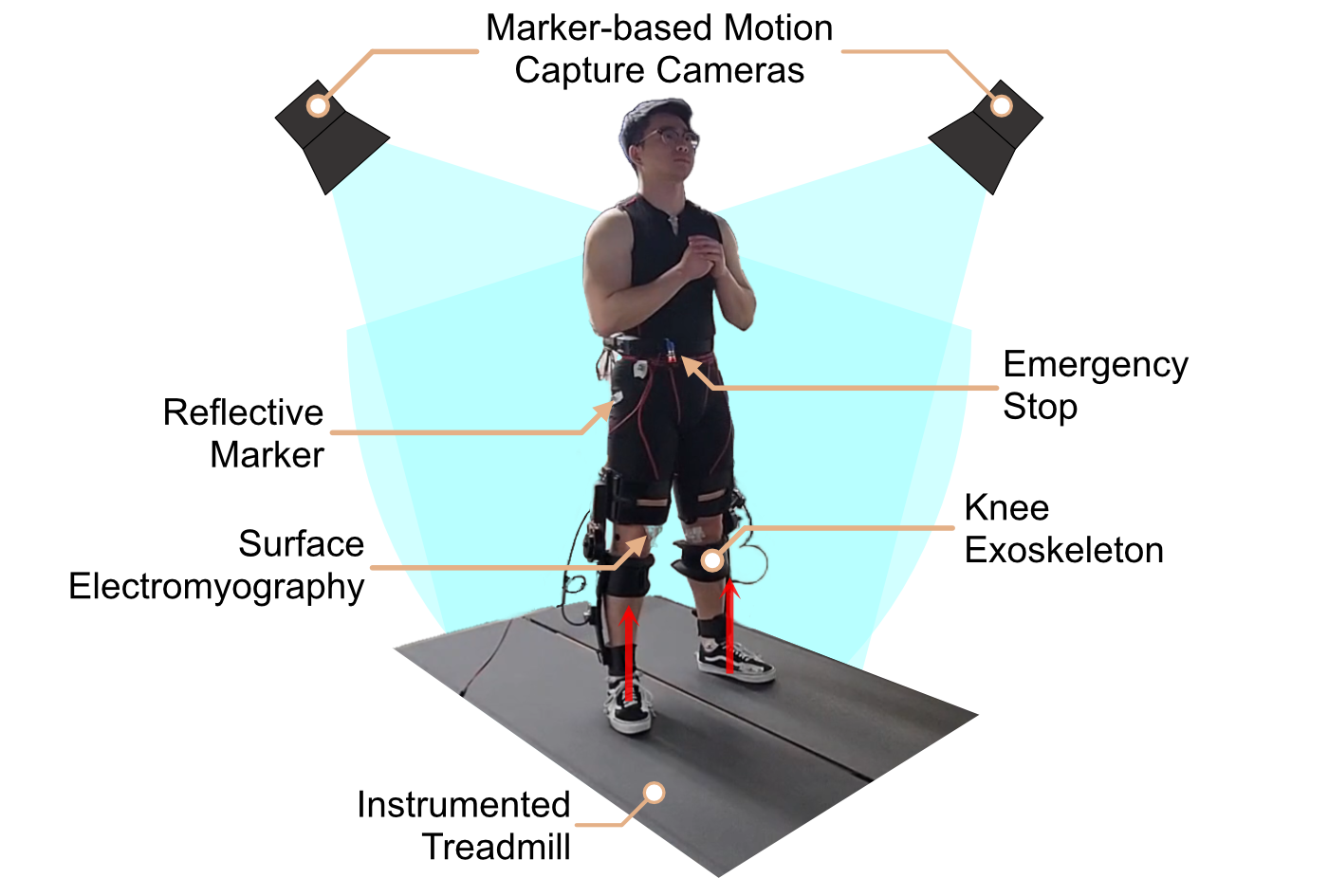}
    \caption{
    \textbf{Experimental setup for validating the integrated exoskeleton and controller during squatting.} A participant wearing the exoskeleton on the split-belt instrumented treadmill, whose embedded force plates measure ground reaction forces, with example placements of the electromyography sensors and reflective motion capture markers.
    }
    \label{fig:Lab_Experiment}
\end{figure}

Seven healthy male participants (age: $26.3 \pm 2.6$ years; body mass: $76.5 \pm 8.7$~kg; height: $1.77 \pm 0.04$~m) were recruited under a protocol approved by the Carnegie Mellon University Institutional Review Board (Protocol STUDY2026\_00000339), and all participants provided written informed consent before the experiment. Data were collected in a motion capture laboratory equipped with a 10-camera system (Vero v2.2, Vicon, UK) and a split-belt instrumented treadmill with two embedded force plates (FIT5, Bertec, USA) (Fig.~\ref{fig:Lab_Experiment}). A total of 24 markers were placed at anatomical landmarks and joint centers for full-body tracking and inverse kinematics in OpenSim \cite{Delp2007OpenSim}, and marker trajectories and ground reaction forces were sampled at 100~Hz. Surface electromyography (EMG) was recorded from the right leg (Trigno Quattro Research+, Delsys, USA) over the vastus medialis, selected as a primary contributor to knee extension torque during squatting, and the medial gastrocnemius, a biarticular muscle that also flexes the knee, monitored to detect antagonist co-contraction or other changes in an unassisted muscle crossing the knee. The hip extensors were not instrumented because the device covered their electrode sites, and sensors placed beneath the cuffs were uncomfortable and disturbed by the applied torque.

After electrode placement, three maximum voluntary contractions (MVCs) were recorded for each muscle, using isometric knee extension against resistance for the vastus medialis and resisted plantarflexion for the medial gastrocnemius. A static calibration trial was then recorded to scale a subject-specific musculoskeletal model. Each participant first completed a 60~s squatting trial without the exoskeleton, with one foot centered on each force plate, paced to a metronome at 40 beats per minute, with each beat marking a change in squat direction. During a subsequent 5-minute rest, the bilateral exoskeleton was donned, its alignment with the knee joint was verified, and the motor encoder zero was set with the knees in full extension. After 5 to 8 familiarization squats and a further 5-minute rest, the assisted squatting trial was performed under a protocol identical to the trial without the exoskeleton.

\begin{figure*}[!t]
    \includegraphics[width=1\textwidth]{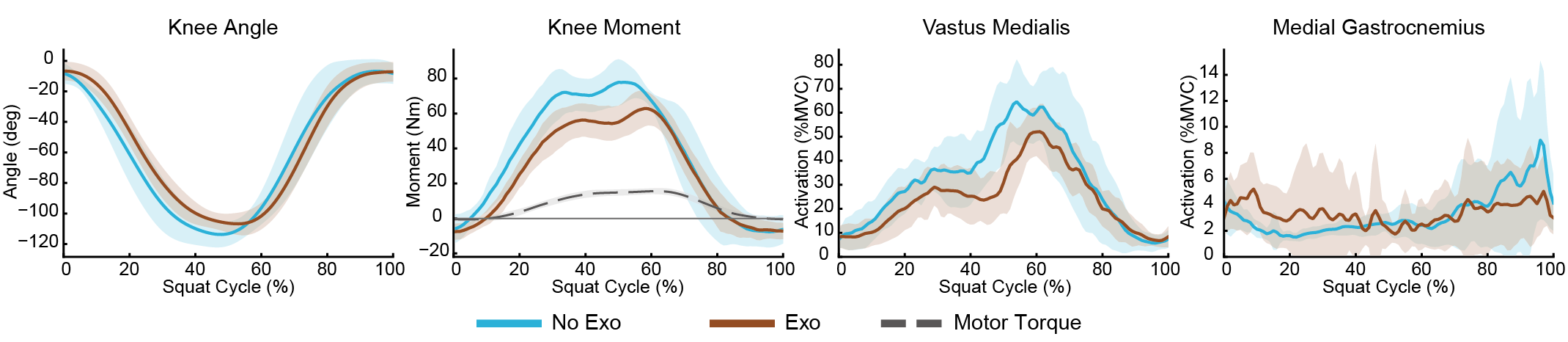}
    \caption{
    \textbf{Group-averaged knee kinematics, kinetics, and muscle activation during squatting.} (a) Knee angle, (b) biological knee extension moment with commanded torque, (c) vastus medialis activation, and (d) medial gastrocnemius activation over the normalized squat cycle. Solid curves represent group means across participants, and shaded regions represent $\pm$1 standard deviation. Blue and brown indicate squatting without (No Exo) and with (Exo) the exoskeleton, respectively, and the dashed gray curve indicates the commanded torque with the exoskeleton. Muscle activation is expressed as a percentage of each participant's maximum voluntary contraction (\%MVC).
       }
    \label{ProfileFig}
\end{figure*}

\begin{figure}[!t]
    \includegraphics[width=1\columnwidth]{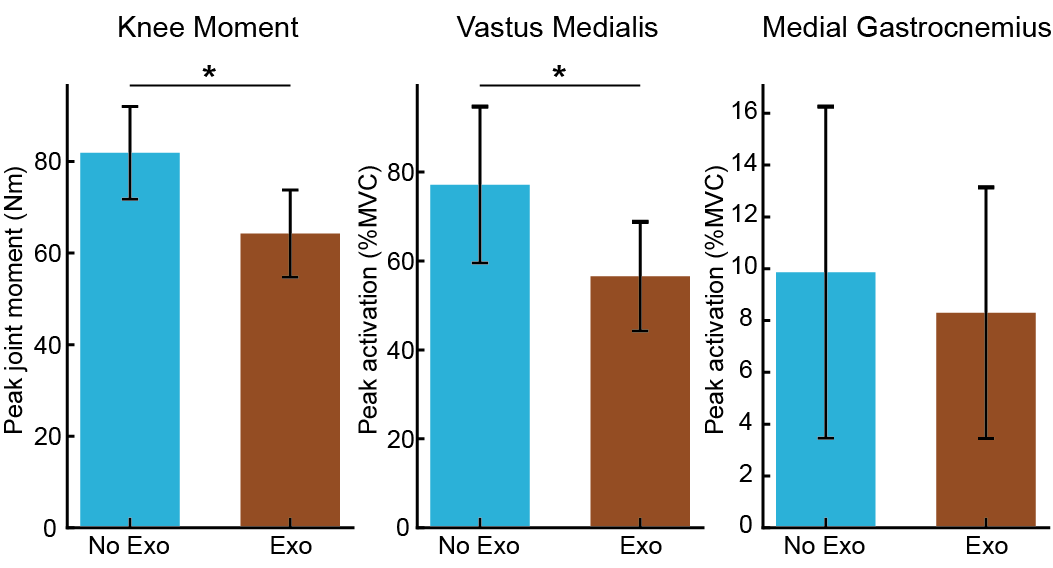}
    \caption{
    \textbf{Peak biological knee extension moment and muscle activation with and without the exoskeleton.} (a) Peak biological knee extension moment, (b) peak vastus medialis activation, and (c) peak medial gastrocnemius activation without (No Exo) and with (Exo) the exoskeleton. Bars represent group means, and error bars represent $\pm$1 standard deviation across participants. Asterisks indicate a significant difference between conditions ($p<0.05$, paired $t$-test).
    }
    \label{BarFig}
\end{figure}

\subsection{Data Processing and Statistical Analysis}
Kinematic and kinetic data were processed in OpenSim \cite{Delp2007OpenSim} with participant-scaled versions of a modified Rajagopal musculoskeletal model \cite{Uhlrich2022MuscleCoordination}. Trials without the exoskeleton were analyzed with the scaled model, whereas trials with the exoskeleton were analyzed with a model augmented to represent the device, with its 3.5~kg mass distributed across the pelvis (0.90~kg), each thigh (0.91~kg), and each shank (0.39~kg), and each segment's diagonal moments of inertia scaled in proportion to its added mass. Inverse kinematics results were low-pass filtered at 6~Hz with a third-order Butterworth filter before inverse dynamics was computed with the condition-specific model. Exoskeleton and motion capture data were aligned using trigger pulses sent from the embedded computer to the motion capture system at the start and end of each trial. In trials with the exoskeleton, the aligned torque command was subtracted from the inverse-dynamics knee moment to obtain the biological knee moment. Individual squat cycles were identified from the vertical pelvis trajectory after fourth-order, zero-phase Butterworth low-pass filtering at 6~Hz, and the final 10 complete cycles of each trial, excluding the initial adaptation period, were time-normalized and averaged within each participant.

EMG signals, recorded at 2148~Hz, were resampled to 1000~Hz, demeaned, band-pass filtered at 10 to 499Hz, full-wave rectified, and low-pass filtered at 8~Hz to obtain linear envelopes, all with fourth-order, zero-phase Butterworth filters. MVC recordings were processed identically, except that values within the upper 1\% of the rectified signal were replaced with neighboring values to limit the influence of transient artifacts. The MVC reference for each muscle was defined as the maximum of a 250~ms moving average of the processed MVC envelope, and EMG signals were normalized to this reference and expressed as a percentage of MVC (\%MVC). The normalized signals were segmented using the squat-cycle boundaries from the kinematic data, time-normalized, and averaged over the same 10 cycles for each participant and condition. Peak knee flexion angle, peak biological knee extension moment, and peak vastus medialis and medial gastrocnemius activation were then extracted from each participant's mean squat-cycle profile and compared between the conditions with and without the exoskeleton using two-sided paired $t$-tests, with statistical significance defined as $p < 0.05$. Data are reported as mean $\pm$ standard deviation.


\section{Results}
\subsection{Knee Joint Kinematics and Kinetics}
We compared knee kinematics and kinetics between the conditions with and without the exoskeleton over the normalized squat cycle. The knee angle profiles followed a similar flexion pattern in both conditions (Fig.~\ref{ProfileFig}), and peak knee flexion decreased from $114.5 \pm 7.7^{\circ}$ without the exoskeleton to $107.1 \pm 7.4^{\circ}$ with the exoskeleton (paired $t$-test, $p = 0.019$). The average biological knee moment profile was lower in magnitude with the exoskeleton (Fig.~\ref{ProfileFig}). Peak biological knee extension moment decreased by 21.5\%, from $81.9 \pm 10.1$~Nm to $64.3 \pm 9.5$~Nm (paired $t$-test, $p = 0.006$; Fig.~\ref{BarFig}), and decreased in all seven participants. With the exoskeleton, peak commanded torque was $15.7 \pm 1.7$~Nm and occurred at $60.1 \pm 5.6$\% of the squat cycle (Fig.~\ref{ProfileFig}).

\subsection{Muscle Activation}
We next compared vastus medialis and medial gastrocnemius activation between the conditions with and without the exoskeleton. The group-averaged vastus medialis activation profile was lower with the exoskeleton (Fig.~\ref{ProfileFig}). Peak vastus medialis activation decreased by 26.7\%, from $77.1 \pm 17.6$\%MVC to $56.5 \pm 12.3$\%MVC (paired $t$-test, $p = 0.016$; Fig.~\ref{BarFig}), and decreased in six of the seven participants. In contrast, medial gastrocnemius activation showed greater variability between participants (Fig.~\ref{ProfileFig}). Peak medial gastrocnemius activation was $9.86 \pm 6.40$\%MVC without the exoskeleton and $8.29 \pm 4.85$\%MVC with the exoskeleton, with no significant difference between conditions (Fig.~\ref{BarFig}).


\begin{table*}[!t]
    \caption{Comparison with Existing Knee Exoskeleton Platforms\label{tab:comparison}}
    \centering
    \footnotesize
    \renewcommand{\arraystretch}{1.2}
    \setlength{\tabcolsep}{3pt}
    \begin{tabular*}{\textwidth}{@{\extracolsep{\fill}} >{\raggedright\arraybackslash}m{2.35cm} >{\raggedright\arraybackslash}m{1.8cm} >{\centering\arraybackslash}m{1.4cm} >{\centering\arraybackslash}m{1.6cm} >{\raggedright\arraybackslash}m{4.25cm} >{\raggedright\arraybackslash}m{4.55cm} @{}}
    \hline\noalign{\vskip 2pt}
    \multicolumn{1}{@{}>{\raggedright\arraybackslash}m{2.35cm}}{Device} &
    \multicolumn{1}{>{\raggedright\arraybackslash}m{1.8cm}}{Validation Subjects} &
    \multicolumn{1}{>{\centering\arraybackslash}m{1.4cm}}{Device Mass (kg)} &
    \multicolumn{1}{>{\centering\arraybackslash}m{1.6cm}}{Peak Torque (Nm)} &
    \multicolumn{1}{>{\raggedright\arraybackslash}m{4.25cm}}{Tasks} &
    \multicolumn{1}{>{\raggedright\arraybackslash}m{4.55cm}@{}}{Reported Reduction}\\[2pt]
    \hline\noalign{\vskip 1pt}
    \textbf{This Study} & 7 able-bodied & 3.5 & 18 & Squatting & 26.7\% vastus medialis activation\\
    Nesler et al. \cite{Nesler2022MBLUE} & 1 able-bodied & 1.6$^{a}$ & 30 & Lifting and lowering & 78.9\% vastus medialis activation\\
    Lee et al. \cite{Lee2024AIUniversalExo} & 10 able-bodied & 4.4 & 15 & Level walking, ramps, and stairs & 6.2\% metabolic cost$^{c}$\\
    Zhang et al. \cite{Zhang2024DualDriveKnee} & 6 able-bodied & 3.2 & 18 & Level walking, ramp and stair ascent & 4.3--12.0\% metabolic cost\\
    Boon et al. \cite{Boon2025StairAscent} & 8 able-bodied & 2.6$^{a}$ & 55 & Stair ascent & 26\% vastus medialis activation\\
    Chen et al. \cite{Chen2021WearableKnee} & 2 able-bodied & 4.1 & 35 & Kneeling & 18.4\% vastus medialis activation$^{b,c}$\\
    Huang et al. \cite{Huang2022StiffnessBased} & 8 able-bodied & 3.5 & 14 & Level walking & 19.6\% vastus medialis activation$^{c}$\\
    Sarkisian et al. \cite{Sarkisian2022SitToStand} & 1 post-stroke & 3.6$^{a}$ & 40 & Sit-to-stand & 29.3\% vastus medialis activation\\
    Yu et al. \cite{Yu2025ControllingPower} & 7 able-bodied & 3.5 & 54 & Squatting & 38.1\% vastus medialis activation,\newline 12.8\% metabolic cost\\
    \hline
    \end{tabular*}
    \par\vspace{2pt}
    \begin{minipage}{\textwidth}
    \scriptsize\raggedright
    Mass is for a bilateral device unless noted. Activation reductions are in peak values and relative to no exoskeleton unless noted. $^{a}$Unilateral device. $^{b}$Mean rather than peak activation. $^{c}$Relative to the unpowered or zero-torque exoskeleton.
    \end{minipage}
\end{table*}

\section{Discussion}
A biological torque controller derived entirely from simulation reduced peak biological knee extension moment and vastus medialis activation during squatting, which supports our hypothesis. Unlike prior biological torque controllers, which were trained on data recorded with the device that runs them \cite{Molinaro2024JointMoment, Molinaro2024TaskAgnostic}, this estimator was learned from open biomechanics data through musculoskeletal motion imitation, and no data from the participants or the device entered training. Across participants, the controller produced consistent assistance, with commanded torque peaking at $15.7 \pm 1.7$~Nm at $60.1 \pm 5.6$\% of the squat cycle, around maximum flexion and the start of ascent, where knee extensor demand is highest \cite{Escamilla2001SquatBiomechanics}. Only the gain and delay were adjusted for each participant during familiarization. Because the delay was tuned for each participant, the timing of peak torque reflects both the estimator output and this tuning, and the present data cannot separate the two. Together, these results indicate that simulation can supply the training data for biological torque control on physical knee exoskeleton hardware during squatting.

The biomechanical measurements are consistent with this assistance. Peak biological knee moment decreased by 17.6~Nm, close to the peak commanded torque, which suggests that the device took over part of the biological effort instead of being counteracted by a larger net joint moment. Because the biological moment was derived using the commanded torque, the 26.7\% reduction in vastus medialis activation provides independent support for this interpretation. The absence of a significant change in medial gastrocnemius activation further indicates that unloading the knee did not detectably transfer effort to the ankle plantarflexors. The 7.4$^{\circ}$ reduction in peak knee flexion indicates a smaller squat depth with the exoskeleton, which may account for part of the moment reduction.

The controller's sensing requirements also shaped the hardware. Because the estimator depends solely on onboard IMU and encoder signals, the cuff rotation observed in the earlier prototype (Section~II-A) corrupted its inputs and diminished the commanded torque. Compliance and migration at the physical interface are known to change how wearable devices transmit assistance \cite{Yandell2017Interface}, and in this exoskeleton they additionally degraded the controller's estimate of the user's state. Because the structural parts could be fabricated in-house, revised cuffs could be produced and tested with the same controller without retraining, and the interface was refined until it met the controller's sensing requirements.

Table~\ref{tab:comparison} places these results among existing knee exoskeletons. Our exoskeleton applies less torque than most of these devices, yet its vastus medialis reduction matches that of a device capable of 55~Nm during stair ascent \cite{Boon2025StairAscent} and trails the 38.1\% reported at 54~Nm during squatting \cite{Yu2025ControllingPower}, where moderate assistance also outperformed maximal assistance. Its distinguishing feature is therefore not peak performance but the combination of accessible hardware with a controller that requires no device data. In related fields, shared hardware platforms that integrate the mechanical design, electronics, and control software have allowed controllers developed by different groups to be compared on the same device \cite{Azocar2020OSL, Williams2025OpenExo}. Our work extends this approach to knee exoskeletons by releasing the hardware together with a training pipeline that derives the controller without extensive data collection.

Several limitations constrain these conclusions. First, only squatting was tested, and whether the estimator, retrained on new reference motions, carries over to tasks with different inter-joint coordination, such as walking or sit-to-stand, remains to be shown. Second, the biological moment was derived from commanded rather than measured torque, so any transmission or interface losses would inflate the estimated assistive contribution. Third, without an unpowered condition, the reported reductions also include the cost of wearing the device and likely understate the effect of assistance alone \cite{Huang2022StiffnessBased}, and the fixed condition order may have introduced fatigue or learning effects. Fourth, torque was capped at 18~Nm by the stock motor driver, and all seven participants were healthy young adults, so the results may not extend to higher assistance levels or to clinical populations \cite{Siviy2023Opportunities}. With validation across these tasks and populations, the exoskeleton and its training pipeline could serve as a shared testbed on which knee exoskeleton controllers are derived from open-source datasets \cite{Scherpereel2023Dataset, Camargo2021Dataset} and compared, helping distinguish the contribution of a controller from that of the hardware it runs on.

\section{Conclusion}
This work presented a knee exoskeleton designed for accessible fabrication and a biological torque controller learned entirely from open-source biomechanics data through musculoskeletal simulation. The controller was deployed on the physical device without any training data collected on the device, with only the assistance gain and delay set for each participant. In seven healthy adults, it reduced peak biological knee extension moment and vastus medialis activation relative to squatting without the exoskeleton. While limited to a single task in this proof-of-concept study, these results demonstrate that simulation can supply the training data for biological torque control on physical exoskeleton hardware. By releasing the hardware together with the training pipeline, this work provides a shared starting point for developing and comparing knee exoskeleton controllers without extensive data collection.

\section*{Acknowledgment}
The authors would like to thank Yunliang Zhao, Min Seo Kim, and Ray Sarabu for their contributions to this work. The authors also thank MetaMobility Lab for the resources for manufacturing, experimentation, and validation of the exoskeleton platform.

\bibliographystyle{IEEEtran}
\bibliography{Knee_Exoskeleton_RA-L}

\end{document}